\documentclass[11pt]{article}
\usepackage{coling2020}
\usepackage{times}
\usepackage{url}
\usepackage{amsmath}
\usepackage{amsfonts}
\usepackage{hhline}
\usepackage{latexsym}
\usepackage{multirow}
\usepackage{graphics}
\usepackage{diagbox}
\usepackage[table]{xcolor}

\title{Attention Word Embedding}
\author{Shashank Sonkar \\
  Rice University \\
  {\tt ss164@rice.edu} \\
  \And
  Andrew E. Waters \\
  Rice University \\
  {\tt aew2@rice.edu} \\\And
  Richard G. Baraniuk \\
  Rice University \\
  {\tt richb@rice.edu} \\
  }
\date{}

\begin{document}
\maketitle

\begin{abstract}
Word embedding models learn semantically rich vector representations of words and are widely used to initialize natural processing language (NLP) models. The popular continuous bag-of-words (CBOW) model of word2vec learns a vector embedding by masking a given word in a sentence and then using the other words as a context to predict it. 
A limitation of CBOW is that it equally weights the context words when making a prediction, which is inefficient, since some words have higher predictive value than others. 
We tackle this inefficiency by introducing the {\em Attention Word Embedding} (AWE) model, which integrates the attention mechanism into the CBOW model. 
We also propose AWE-S, which incorporates subword information. 
We demonstrate that AWE and AWE-S outperform the state-of-the-art word embedding models both on a variety of word similarity datasets and when used for initialization of NLP models.

\end{abstract}

\section{Introduction}
Word embedding models learn vector representations of words such that words that are semantically related are close to each other in the vector space. Popular word embedding models include word2vec \cite{mikolov2013efficient,mikolov2013distributed}, GloVe \cite{pennington2014glove}, and fastText \cite{bojanowski2017enriching}. Word embedding models are often used to initialize deep learning models for various natural language processing (NLP) tasks such as machine translation \cite{bahdanau2014neural}, part-of-speech tagging \cite{ling2015two}, and sentiment analysis \cite{wang2016recursive}, leading to improved performance over training the models from scratch.

The core idea underlying the training methodology of all the above embedding models is that ``a word is characterized by the company it keeps" \cite{firth1957synopsis}. For instance, the continuous bag-of-words (CBOW) model of word2vec predicts a randomly selected word in a sentence using the other words in the sentence as a context.
The context words are typically treated equally by these models. 
However, clearly some words in the context will be more predictive of the masked word than others, and intuitively, should be weighted more heavily.

A promising remedy that treats each context word differently is the {\it attention mechanism} \cite{vaswani2017attention}, which enables learning the relative importance of input features for the prediction of the desired output. 
Attention has become the de facto standard and state-of-the-art in modern NLP for a range of different tasks \cite{zhang2018self,devlin2018bert,dai2019transformer}.
However, to date, there have been only a few attempts to employ it to improve the performance and interpretability of word embedding models \cite{ling2015not,schick2019attentive}.

\subsection{Contributions}

In this paper, we propose the {\em Attention Word Embedding} (AWE), a new word embedding model that integrates the attention mechanism into the CBOW model of word2vec.  

AWE consists of two components. First, AWE uses a variant of self-attention \cite{vaswani2017attention} to narrow down relevant words in the context for the prediction of the masked word. In contrast to prior work \cite{ling2015not,schick2019attentive}, the attention weights are a function of both the context words and the target/masked word. Second, AWE embeds the context words and the masked word representations in a shared subspace, since both representations embody the meaning of the word. Note that this is in sharp contrast with CBOW, which learns two different embeddings for each word, one when the word is present in the context and another when it is masked.  

We also introduce a variant of AWE, AWE-S, which leverages subword information to enrich the word vectors. The idea to use subwords is inspired from fastText \cite{bojanowski2017enriching}, which defines subwords of a word as its character n-grams. Incorporating subword information is particularly valuable for improving word vector representations for rare and unseen words.

Our experimental evaluations show the superior performance of AWE and AWE-S than many existing word embedding models in both language modeling tasks and a number of downstream NLP applications including natural language inference, sentence semantic relatedness, and paraphrase detection. Furthermore, we analyze and interpret the role of the attention weights in AWE, which provide interesting insights into the workings of the attention mechanism.

\section{Preliminaries}
\subsection{Word2Vec}
Word2vec is one of the standard word embedding models \cite{mikolov2013efficient,mikolov2013distributed}. There are two architectures proposed for word2vec: CBOW and Skip-Gram. We focus on the CBOW model in this work.

CBOW predicts a masked word using its context (\textit{fill in the blanks} model). For each word, it learns two vectors to represent the two roles that a word can perform - first, when it is present in the context of the masked word, and second, when it is masked.

Let $U = [\boldsymbol{u}_1,...,\boldsymbol{u}_N]^T \in \mathbb{R}^{N \times D}$ and $V =  [\boldsymbol{v}_1,...,\boldsymbol{v}_N]^T \in \mathbb{R}^{N \times D}$ where $N$ is the size of the vocabulary, and $D$ is the size of the word vector. $U$ models the first role and is used to calculate the context vector, $\boldsymbol{c}$, given by
\begin{equation}
    \boldsymbol{c} = \sum_{i \in [-b, b] - \{0\}} \boldsymbol{u}_{w_i},
    \label{eq:context_vector}
\end{equation}
where $b$ is the size of the context window and $w_i$ is the index of each word ($w_0$ is the index of the masked word; the rest are the indices of the context words). $V$ models the second role and learns the masked word vector for $w_0$, given by $\boldsymbol{v}_{w_0}$. The probability $p$ of $w_0$ to occur in the context of $\{w_{-b},.., w_{-1}, w_{1},.., w_{b}\}$ is given by
\begin{equation}
    p(w_0|W_{[-b, b] - \{0\}}) \propto \exp{\boldsymbol{v}_{w_0}^T \boldsymbol{c}}.
    \label{eq:prob}
\end{equation}
    
The Skip-Gram model of word2vec is similar to CBOW. The key difference is that it predicts the context words using the masked word.

\section{New Model: Attention Word Embedding (AWE)}

\subsection{The AWE Model}

We now explain AWE, our proposed word embedding model that incorporates the attention mechanism.
AWE augments the CBOW model of word2vec with the attention mechanism in two different ways. First, we introduce two new matrices, a key matrix, $K \in \mathbb{R}^{N \times D'}$ and a query matrix, $Q \in \mathbb{R}^{N \times D'}$, where $N$ is the size of the vocabulary and $D' \in \mathbb{Z}$. With the attention mechanism, the context vector $\boldsymbol{c}$ is not modeled as a simple sum in~(\ref{eq:context_vector}) but rather as a {\it weighted} sum of context word vector embeddings
\begin{equation*}
    \boldsymbol{c_{\rm attn}} = \sum_{i \in [-b, b] - \{0\}} a_{w_i}\boldsymbol{u}_{w_i}\,,
\end{equation*}
where $a_{w_i}$ is the attention weight of each context word vector $\boldsymbol{u}_{w_i}$ 
calculated using the key matrix $K$ and the query matrix $Q$
\begin{equation*}
    a_{w_i} =
    \exp\left({\boldsymbol{k_{w_0}}^T\boldsymbol{q_{w_i}}}\right)\,,
\label{eq:attention}
\end{equation*}

Second, we share the weights between the context word embedding matrix and the masked word embedding matrix in our model, i.e., we set $U = V$. Sharing weights is a natural and intuitive choice, since both matrices embed the meaning of a word in its vector representation, and the meaning of the word remains the same irrespective of it occurring in the context window, or as the masked word. In CBOW, the choice is to have two separate matrices, $U$ and $V$, is justified as it leads to increase in performance. However, in AWE, the intuitive choice to have $U$ same as $V$ works better and adds interpretability to the model. On top of that, it has an added advantage. The number of parameters in AWE are much less as compared to CBOW even though AWE has one more matrix than CBOW. The reason is that the key matrix $K \in \mathbb{R}^{N \times D'}$ and the query matrix $Q \in \mathbb{R}^{N \times D'}$ are much smaller as compared to the value matrix, $V = [\boldsymbol{u}_1, ..., \boldsymbol{u}_D]^T \in \mathbb{R}^{N \times D}$. In our experiments, we set $D'=50$  and $D=500$.

\subsection{AWE-S: A Subword Variant of AWE}
We also propose AWE-S, a variant of AWE that leverages {\it subword} information. A subword of a word is a part of the word, e.g., its character n-grams or its lemmas. Subword information has been used in a number of word embedding models including fastText \cite{bojanowski2017enriching} and BERT \cite{devlin2018bert} to improve the expressiveness of rare word representations. As an illustration, learning a robust representation for a word like \textit{awing}, which occurs rarely in a corpus, is challenging due to a dearth of examples. To circumvent the scarcity of samples, a word embedding model can exploit the fact that \textit{awe} is the verb lemma form of \textit{awing} to learn a better embedding for \textit{awing}.

In AWE-S, the embedding of each word is given by the sum of embeddings of all of its subwords, regardless of whether the word is a context word or the masked word in the prediction task, i.e.,
\begin{equation*}
    \boldsymbol{\tilde{u}_{w}} = \sum_{e_j \in  S_w}\boldsymbol{u}_{e_j}\,,
\end{equation*}
where $S_w$ is the subword set of the word $w$.
The context vector $\boldsymbol{c}$ in AWE-S is given by
\begin{equation*}
    \boldsymbol{c_{\rm attn}} = \sum_{i \in [-b, b] - \{0\}} a_{w_i} \left( \sum_{e_j \in  S_{w_i}}\boldsymbol{\tilde{u}}_{e_j}\right)\,,
\end{equation*}
and the probability of the masked word $w_0$ to occur in the context of $\{w_{-b},.., w_{-1}, w_{1},.., w_{b}\}$ is given by
\begin{equation}
    p(w_0|W_{[-b, b] - \{0\}}) \propto \exp{\boldsymbol{\tilde{u}_{w_0}}^T \boldsymbol{c_{\rm attn}}}\,.
    \label{eq:prob_awe_fast}
\end{equation}
Note that in fastText, the context vector of \eqref{eq:context_vector} and the probability of $w_0$ to occur in the context of $\{w_{-b},.., w_{-1}, w_{1},.., w_{b}\}$  is given by
\begin{equation}
    \boldsymbol{c} = \sum_{i \in [-b, b] - \{0\}} \sum_{e_j \in  S_{w_i}}\boldsymbol{u}_{e_j}\,,\qquad
    p(w_0|W_{[-b, b] - \{0\}}) \propto \exp{\boldsymbol{v}_{w_0}^T \boldsymbol{c}}\,.
    \label{eq:prob_fast}
\end{equation}

The computations in AWE-S may appear to be similar those in fastText.
However, observe that, in fastText, a word is not represented as the sum of the embeddings of its subwords when it is masked. Also, contrary to fastText, which uses character n-grams to construct the subword set for a word, AWE-S uses the noun, adjective, and verb lemmas of the word to construct the subword set. This limits the size of the subword set significantly as compared to using character n-grams, which helps to reduce the number of learnable parameters in AWE-S.

\begin{table*}[t]
\centering
\resizebox{\columnwidth}{!}{
\begin{tabular}{|c|c|c|c|c|c|c|c|c|} \hline
\textbf{Model} & \textbf{MEN} & \textbf{WS353} & \textbf{WS353R} & \textbf{WS353S} & \textbf{SimLex999} & \textbf{RW(RareWords)} & \textbf{RG65} & \textbf{MTurk} \\ \hline
\textbf{AWE} & \textbf{0.771 }& \textbf{0.672} & \textbf{0.609} & \textbf{0.773} & 0.377 & \textbf{0.440} & \textbf{0.836} & \textbf{0.683} \\ \hline
\textbf{CBOW} & 0.717 & 0.591 & 0.478 & 0.720 & \textbf{0.383} & 0.396 & 0.791 & 0.659 \\ \hline
\textbf{SG} & 0.738 & 0.664 & 0.607 & 0.729 & 0.371 & 0.440 & 0.769 & 0.662 \\ \hline
\textbf{GloVe} & 0.695 & 0.553 & 0.492 & 0.683 & 0.324 & 0.340 & 0.763 & 0.621 \\ \hline
\end{tabular}
}
\caption{\centering Spearman correlation on word similarity datasets for AWE, CBOW, Skip-Gram, and GloVe.}
\label{tab:results_wsd1}
\vspace{5mm}

\centering
\resizebox{\columnwidth}{!}{
\begin{tabular}{|c|c|c|c|c|c|c|c|c|} \hline
\textbf{Model} & \textbf{MEN} & \textbf{WS353} & \textbf{WS353R} & \textbf{WS353S} & \textbf{SimLex999} & \textbf{RW(RareWords)} & \textbf{RG65} & \textbf{MTurk} \\ \hline
\textbf{AWE-S} & \textbf{0.771} & 0.675 & 0.610 & \textbf{0.769} & \textbf{0.386} & \textbf{0.463} & \textbf{0.819} & \textbf{0.692} \\ \hline
\textbf{FastText} & 0.761 & \textbf{0.691} & \textbf{0.642} & 0.745 & 0.385 & 0.457 & 0.795 & 0.685 \\ \hline
\end{tabular}
}
\caption{\centering Spearman correlation on word similarity datasets for FastText and AWE-S, which use additional subword information.}
\label{tab:results_wsd2}
\vspace{5mm}

\centering
\resizebox{\columnwidth}{!}{
\begin{tabular}{|c|c|c|c|c|c|}
\hline
\textbf{Model} & \textbf{SNLI (Acc)} & \textbf{SICK-E (Acc)} & \textbf{SICK-R (Pearson)} & \textbf{STS Benchmark (Pearson)} & \textbf{MRPC (F1)} \\ \hline
\textbf{AWE} & \textbf{65.23$^\star$} & 76.05$^\star$ & 0.783$^\ast$ & 0.648$^\ast$ & \textbf{81.70$^\dagger$} \\ \hline
\textbf{AWE-S} & 65.21$^\star$ & 76.05$^\star$ & \textbf{0.785$^\ast$} & \textbf{0.651$^\ast$} & 81.49$^\dagger$ \\ \hline
\textbf{CBOW} & 65.07$^\star$ & \textbf{76.23$^\star$} & 0.781$^\ast$ & 0.600$^\ast$ & 81.26$^\dagger$ \\ \hline
\textbf{SG} & 64.45$^\star$ & 75.04$^\star$ & 0.780$^\ast$ & 0.632$^\ast$ & 81.12$^\dagger$ \\ \hline
\textbf{FastText} & 64.98$^\star$ & 75.06$^\star$ & 0.782$^\ast$ & 0.586$^\ast$ & 80.88$^\dagger$ \\ \hline
\textbf{GloVe} & 64.18$^\star$ & 75.91$^\star$ & 0.782$^\ast$ & 0.643$^\ast$ & 80.60$^\dagger$ \\ \hline
\end{tabular}
}
\caption{\centering Evaluation results of AWE and its subword variant, AWE-S, against other word embedding methods for initializing models for downstream tasks. Accuracy is reported for SNLI dataset and SICK-E, Pearson correlation is reported for SICK-R and STS Benchmark, and F1 score is reported for MRPC. `$^\star$', `$^\ast$' and  `$^\dagger$' denote accuracy, pearson correlation, and F1 score respectively.}
\label{tab:results_downstream_tasks}
\end{table*}

\section{Experiments}

We demonstrate the superior performance of AWE and AWE-S against CBOW, Skip-Gram, GloVe, and fastText on a variety of datasets, which broadly fall into two main categories:
\begin{enumerate}
    \item \textbf{Word similarity tasks.} We use this task to evaluate the performance of word embedding models themselves. The datasets for this task contain word pairs and a semantic similarity score associated with the pairs. The scores were annotated by human subjects.  These are the most widely used evaluation methods for word embedding models \cite{bakarov2018survey}. We test AWE's performance using spearman correlation score on eight such datasets - MEN \cite{bruni2014multimodal}, WS353 \cite{finkelstein2001placing}, WS353R \cite{agirre2009study}, WS353S \cite{agirre2009study}, SimLex999 \cite{hill2015simlex}, RW(RareWords) \cite{luong2013better}, RG65 \cite{rubenstein1965contextual}, and MTurk \cite{radinsky2011word}.
    
    \item \textbf{Downstream NLP tasks.} These tasks evaluate performance on important NLP applications including natural language inference, semantic entailment, semantic relatedness, and paraphrase detection. 
    Because the conventional application of word embedding models has been to initialize NLP models,
    we use these tasks to assess the quality of our word embeddings for initializing NLP models for downstream applications.  
    Datasets include SNLI \cite{bowman2015large}, SICK-E \cite{marelli2014sick}, SICK-R \cite{marelli2014sick}, STS Benchmark \cite{cer2017semeval}, and MRPC \cite{dolan2004unsupervised}.
    Unlike datasets for the previous task, all the datasets here have a training set.
    
\end{enumerate}
 
\subsection{Experimental Setup}
We train AWE, CBOW, Skip-Gram, GloVe, and fastText on the wikipedia dataset ($\sim 17$  GB in size), with vocabulary consisting of one million words. Dimensions of word embeddings for CBOW, Skip-Gram, GloVe, and fastText is 500. For AWE and AWE-S, $K, Q \in \mathbb{R}^{N \times 50}$, and $V \in \mathbb{R}^{N \times 500}$, where $N$ is the size of the vocabulary. Widely-used parameter settings (context window size is 5, number of negative samples is 5, number of training epochs is 5) are used to train CBOW, Skip-Gram, and fastText. AWE and AWE-S are also trained for the same number of epochs. GloVe is trained for 100 epochs and parameters are set to those recommended in the paper (x-max is 100 and context window size is 10). We used open-source word embedding evaluation tool kits for assessing the quality of the models \cite{jastrzebski2017evaluate,conneau2018senteval}.

\subsection{Numerical Results}
We report the performance statistics of AWE, AWE-S, and the competing word embedding models on word similarity datasets and downstream NLP tasks in Tables~\ref{tab:results_wsd1}, \ref{tab:results_wsd2}, and \ref{tab:results_downstream_tasks}.

Table~\ref{tab:results_wsd1} shows performance of AWE, CBOW, Skip-Gram, and GloVe on word similarity datasets. We use spearman correlation between cosine similarity and human-annotated relatedness scores as the metric to measure the performance. AWE performs significantly better on all of the datasets except SimLex999.

Table~\ref{tab:results_wsd2} compares performance of AWE-S vs fastText. Both algorithms incorporate additional subword information into their architecture. FastText uses character n-grams, while AWE-S uses the lemma forms of the words. AWE wins against FastText on six out of the eight datasets.

Table~\ref{tab:results_downstream_tasks} reports performance of AWE and AWE-S against CBOW, Skip-Gram, GloVe, and fastText on supervised downstream tasks. A logistic regression classifier is trained to classify sentences using pre-trained word embeddings. For these initialization applications, AWE provides improvement in performance as compared to other models over all datasets except SICK-E.

\begin{table*}[t]
\centering
\resizebox{0.95\columnwidth}{!}{
\begin{tabular}{|c|c|c|c|c|c|c|c|}
\hline
\textbf{\backslashbox{Mask}{Sentence}} & & professor & scolded & students & \cellcolor{gray!55} for & playing & games \\ \hline
\multirow{2}{*}{ professor } & attention weight & - & 1.112 & \cellcolor{gray!20} \textbf{1.666} & 29.096 & 1.125 & 0.171 \\ \cline{2-8}
& word vector similarity & - & 0.005 & 0.381 & -0.003 & 0.037 & -0.228 \\ \hline
\multirow{2}{*}{ playing } & attention weight & 1.392 & 0.932 & 0.641 & 2190.342 & - & \cellcolor{gray!20} \textbf{2.278} \\ \cline{2-8}
& word vector similarity & 0.037 & 0.048 & 0.116 & 0.001 & - & 0.975 \\ \hline
\end{tabular}
}
\vspace{5mm}

\centering
\resizebox{0.95\columnwidth}{!}{
\begin{tabular}{|c|c|c|c|c|c|c|c|}
\hline
\textbf{\backslashbox{Mask}{Sentence}}  & & mother & \cellcolor{gray!55} and & child & crossing & \cellcolor{gray!55}the & road \\ \hline
\multirow{2}{*}{ child } & attention weight & \cellcolor{gray!20} \textbf{2.527} & 746.334 & - & 0.55 & 0.073 & 0.827 \\ \cline{2-8}
& word vector similarity & 1.076 & 0 & - & -0.017 & -0.005 & -0.22 \\ \hline
\multirow{2}{*}{ road } & attention weight & 0.353 & 728.901 & 0.553 & \cellcolor{gray!20} \textbf{1.86} & 1734.955 & - \\ \cline{2-8}
& word vector similarity & -0.05 & 0 & -0.22 & 0.961 & 0 & - \\ \hline
\end{tabular}
}
\vspace{5mm}

\centering
\resizebox{\columnwidth}{!}{
\begin{tabular}{|c|c|c|c|c|c|c|c|c|}
\hline
\textbf{\backslashbox{Mask}{Sentence}}  & & my & elementary & school & \cellcolor{gray!55} has & \cellcolor{gray!55} a & skating & rink \\ \hline
\multirow{2}{*}{ skating } & attention weight & 0.243 & 0.434 & \cellcolor{gray!20} \textbf{1.384} & 39.814 & 274.934 & - & \cellcolor{gray!20} \textbf{1.354} \\ \cline{2-9}
& word vector similarity & -0.111 & 0.062 & 0.206 & -0.004 & -0.001 & - & 3.636 \\ \hline
\multirow{2}{*}{ school } & attention weight & 0.138 & \cellcolor{gray!20} \textbf{0.972} & - & 21.477 & 1066.489 & 0.369 & 0.626 \\ \cline{2-9}
& word vector similarity & -0.011 & 2.595 & - & -0.002 & 0 & 0.206 & 0.101 \\ \hline
\end{tabular}
}
\caption{Interpretation of attention. Attention weight between the masked word and the context word is given by $e^{(\boldsymbol{k}_{\text{masked word}})^T\boldsymbol{q}_{\text{context word}}}$, while the word vector similarity between the masked word and the context word is given by  $e^{(\boldsymbol{u}_{\text{masked word}})^T\boldsymbol{u}_{\text{context word}}}$. Highly frequent words are highlighted with a \colorbox{gray!55}{dark gray background}. For each masked word, the attention weights corresponding to context words that are most attended to (excluding highly frequent words) are highlighted with \textbf{bold} numerals and \colorbox{gray!20}{light gray background}.}
\label{tab:analyze_attention_eg}
\end{table*}

\begin{table*}[t]
\centering
\resizebox{\columnwidth}{!}{
\begin{tabular}{|c|c|c|c|c|c|c|c|c|c|c|c|c|}
\hhline{*{13}-}
\textbf{\backslashbox{Mask}{Sentence}} & & colored & leaves & make & autumn & beautiful & & & & & & \\ \hhline{*{13}-}
\multirow{2}{*}{ autumn } & att. wt. & 0.905 & \cellcolor{gray!20} \textbf{2.828} & 0.069 & - & 0.561 & & & & & & \\ \cline{2-13}
& sim & -0.006 & 0.562 & -0.116 & - & 0.157 & & & & & & \\ \hhline{*{13}-}
\multicolumn{13}{|c|}{ } \\ \hhline{*{13}-}
& & soldier & returned & home & after & \cellcolor{gray!55} the & battle & & & & & \\ \hhline{*{13}-}
\multirow{2}{*}{ soldier } & att. wt. & - & 1.593 & 0.861 & 0.044 & 480.195 & \cellcolor{gray!20} \textbf{2.379} & & & & & \\ \cline{2-13}
& sim & - & 0.065 & -0.002 & -0.108 & -0.001 & 0.67 & & & & & \\ \hhline{*{13}-}
\multicolumn{13}{|c|}{ } \\ \hhline{*{13}-}
& & extra & oil & helps & \cellcolor{gray!55} to & add & flavor & \cellcolor{gray!55} to & food & & & \\ \hhline{*{13}-}
\multirow{2}{*}{ oil } & att. wt. & 1.202 & - & 0.446 & 599.004 & 0.587 & \cellcolor{gray!20} \textbf{1.864} & 599.004 & \cellcolor{gray!20} \textbf{2.965} & & & \\ \cline{2-13}
& sim & 0.274 & - & -0.142 & -0.001 & -0.071 & 0.744 & -0.001 & 0.973 & & & \\ \hhline{*{13}-}
\multicolumn{13}{|c|}{ } \\ \hhline{*{13}-}
& & please & open & \cellcolor{gray!55} the & - & \cellcolor{gray!55} to & chapter & \cellcolor{gray!55} for & reading & today & & \\ \hhline{*{13}-}
\multirow{2}{*}{ book } & att. wt. & 1.559 & 0.104 & 2090.525 & & 0.65 & \cellcolor{gray!20} \textbf{2.74} & 31.606 & \cellcolor{gray!20} \textbf{4.858} & 0.42 & & \\ \cline{2-13}
& sim & 0.042 & -0.292 & 0.001 & & -0.004 & 0.952 & -0.004 & 0.529 & 0.088 & & \\ \hhline{*{13}-}
\multicolumn{13}{|c|}{ } \\ \hhline{*{13}-}
& & watch & \cellcolor{gray!55} is & \cellcolor{gray!55} a & small & clock & carried & \cellcolor{gray!55} or & worn & \cellcolor{gray!55} by & \cellcolor{gray!55} a & person \\ \hhline{*{13}-}
\multirow{2}{*}{ watch } & att. wt. & - & 1.977 & 32.681 & 0.388 & \cellcolor{gray!20} \textbf{3.47} & 1.855 & 714.149 & 1.575 & 159.371 & 32.681 & 1.29 \\ \cline{2-13}
& sim & - & -0.005 & -0.004 & -0.188 & 0.921 & 0.223 & 0 & 0.28 & -0.002 & -0.004 & 0.097 \\ \hhline{*{13}-}
\multicolumn{13}{|c|}{ } \\ \hhline{*{13}-}
& & phone & \cellcolor{gray!55} is & \cellcolor{gray!55} a & communication & device & \cellcolor{gray!55} that & permits & users & \cellcolor{gray!55} to & conduct & conversation \\ \hhline{*{13}-}
\multirow{2}{*}{ phone } & att. wt. & - & 5.54 & 755.948 & \cellcolor{gray!20} \textbf{1.735} & \cellcolor{gray!20} \textbf{2.634} & 176.562 & 0.902 & 1.205 & 567.022 & 755.948 & \cellcolor{gray!20} \textbf{4.791} \\ \cline{2-13}
& sim & - & -0.003 & 0 & 0.716 & 1.19 & -0.002 & 0.002 & 0.508 & -0.001 & 0 & 0.869 \\ \hhline{*{13}-}
\end{tabular}
}
\caption{More examples of attention weights between the masked word and the context words. Attention weight (att. wt.) between the masked word and context word is given by $e^{(\boldsymbol{k}_{\text{masked word}})^T\boldsymbol{q}_{\text{context word}}}$, while the word vector similarity (sim.) between the masked word and context word is given by  $e^{(\boldsymbol{u}_{\text{masked word}})^T\boldsymbol{u}_{\text{context word}}}$. Highly frequent words are highlighted with a \colorbox{gray!55}{dark gray background}. For each masked word, the attention weights corresponding to context words that are most attended to (excluding highly frequent words) are highlighted with \textbf{bold} numerals and a \colorbox{gray!20}{light gray background}.}
\label{tab:analyze_attention_eg2}
\end{table*}


\subsection{Interpretation of the attention mechanism}

Studies have shown that the attention mechanism helps a neural network to attend to relevant features in the input \cite{wiegreffe2019attention,vashishth2019attention}.
Thus, motivating the integration of the attention mechanism in CBOW. However, some studies argue otherwise and show that the attention weights are poor indicators of feature importance \cite{jain2019attention,serrano2019attention}. This differing opinion thus necessitates an investigation as to why attention works in the case of AWE. We visually analyze the attention weights to investigate if the attention mechanism models the importance of a context word for the masked word prediction in AWE. The investigation revealed two key findings.

First, no matter the masked word, the attention weight of the masked word and a highly frequent word is typically high. In Table~\ref{tab:analyze_attention_eg} and \ref{tab:analyze_attention_eg2}, the attention weights for highly frequent words like \textit{for, and, the, has,} and \textit{a} (words that have cells with dark gray background) are quite high as compared to other words in the sentence. Note that even though the attention weight is large, the similarity between word vectors is very close to zero, thus not affecting the probability of prediction of the masked word.
    
Second, if we leave out these highly frequent words from the set of context words, we observe that the attention weights focus on more informative words in the context for the prediction of the masked word. In Table~\ref{tab:analyze_attention_eg}, for each masked word, the attention weights corresponding to context words that are most attended to for its prediction (excluding frequent words) has been highlighted with bold numerals and a light gray background. For more examples, please refer to Table~\ref{tab:analyze_attention_eg2}.

In a nutshell, the attention mechanism in AWE focuses on informative context words for the prediction of the masked word if highly frequent words are removed from the set of context words.

\subsection{Role of subword information}
Subwords play a critical function in the robust learning of infrequent words. For instance, the word \textit{happiest} may not occur as frequently as the word \textit{happy}, but the model can learn more about the word \textit{happiest} if it were to supplement it with the information of the word \textit{happy} as well. This is the reason why fastText and AWE-S perform significantly better than other embedding methods on the RareWords dataset (Table~\ref{tab:results_wsd1} and Table~\ref{tab:results_wsd2}). FastText relates the two words \textit{happy} and \textit{happiest} through the n-gram \textit{`happ'}. On the other hand, AWE-S relates the word \textit{happiest} to \textit{happy} through its verb lemma form, i.e, \textit{`happy'}.

\section{Conclusions and Future Work}

In this work, we have proposed AWE and AWE-S, which perform significantly better than CBOW, Skip-Gram, GloVe, and fastText on a variety of datasets across a diverse set of tasks. 
The simple setting of masked word prediction in AWE also helped us visually analyze and interpret how the attention mechanism works. Our analysis revealed that the attention mechanism can figure out the context words that are most relevant for predicting the masked word. Future research will extend the analysis through data-driven methods, which can provide a deeper understanding of the attention mechanism.

\bibliographystyle{acl}
\bibliography{bib_file}

\end{document}